\documentclass[10pt,twocolumn,letterpaper]{article}

 \usepackage[pagenumbers]{cvpr} 

\usepackage[dvipsnames]{xcolor}

\definecolor{cvprblue}{rgb}{0.21,0.49,0.74}
\usepackage[pagebackref,breaklinks,colorlinks,citecolor=cvprblue]{hyperref}

\usepackage{mathrsfs}
\usepackage{graphicx}
\usepackage{subcaption}
\usepackage{multirow}
\usepackage{colortbl}
\usepackage[accsupp]{axessibility}

\def\paperID{3142} 
\def\confName{CVPR}
\def\confYear{2024}

\definecolor{light-gray}{gray}{0.85}

\title{Video-Based Human Pose Regression via Decoupled Space-Time Aggregation}

\author{Jijie He, Wenwu Yang\thanks{Correspondence Author (wwyang@zjgsu.edu.cn)}\\
Zhejiang Gongshang University, China
}

\begin{document}
\maketitle

\renewcommand{\thefootnote}{\dag}

\begin{abstract}
By leveraging temporal dependency in video sequences, multi-frame human pose estimation algorithms have demonstrated remarkable results in complicated situations, such as occlusion, motion blur, and video defocus. These algorithms are predominantly based on heatmaps, resulting in high computation and storage requirements per frame, which limits their flexibility and real-time application in video scenarios, particularly on edge devices. In this paper, we develop an efficient and effective video-based human pose regression method, which bypasses intermediate representations such as heatmaps and instead directly maps the input to the output joint coordinates. Despite the inherent spatial correlation among adjacent joints of the human pose, the temporal trajectory of each individual joint exhibits relative independence. In light of this, we propose a novel Decoupled Space-Time Aggregation network (DSTA) to separately capture the spatial contexts between adjacent joints and the temporal cues of each individual joint, thereby avoiding the conflation of spatiotemporal dimensions. Concretely, DSTA learns a dedicated feature token for each joint to facilitate the modeling of their spatiotemporal dependencies. With the proposed joint-wise local-awareness attention mechanism, our method is capable of efficiently and flexibly utilizing the spatial dependency of adjacent joints and the temporal dependency of each joint itself. Extensive experiments demonstrate the superiority of our method. 
Compared to previous regression-based single-frame human pose estimation methods, DSTA significantly enhances performance, achieving an \textbf{8.9} mAP improvement on PoseTrack2017. 
Furthermore, our approach either surpasses or is on par with the state-of-the-art heatmap-based multi-frame human pose estimation methods. Project page: \href{https://github.com/zgspose/DSTA}{https://github.com/zgspose/DSTA}. 
\end{abstract}

\section{Introduction}
\label{sec:intro}

Human pose estimation, which aims at identifying anatomical keypoints (\textit{e.g.}, elbow, knee, etc.)  of human bodies from images or videos, has been extensively studied in the computer vision community~\cite{TokenPose_CVPR2023,CID_cvpr2022,Vitpose_NIPS2022,HRFormer_NIPS2021}. It plays a crucial role in a variety of human-centric tasks, including motion capture, activity analysis, surveillance, and human-robot interaction~\cite{PoseFormer_ICCV2021}. Recently, significant progress has been made in the field of human pose estimation, particularly with the advent of deep convolutional neural networks (CNNs)~\cite{Resnet_cvpr2016,HRNet_CVPR2019,mobilenetv2_cvpr2018} and Transformer networks~\cite{Attention_NIPS2017,ViT_2021}. 
While the majority of recent methods focus on estimating human poses in \textit{static images}, it has been demonstrated in~\cite{PoseWarper_NIPS2019,DCPose_CVPR2021,TDMI_CVPR2023,} that the significance of \textit{dynamic cues} (\textit{i.e.}, temporal dependency and geometric consistency) across video frames cannot be overlooked. To address inherent challenges in human motion images, such as motion blur, video defocus, and pose occlusions, it is essential to sufficiently exploit the temporal cues in video sequences.

\begin{figure}
    \centering
    \includegraphics[width=1.0\linewidth]{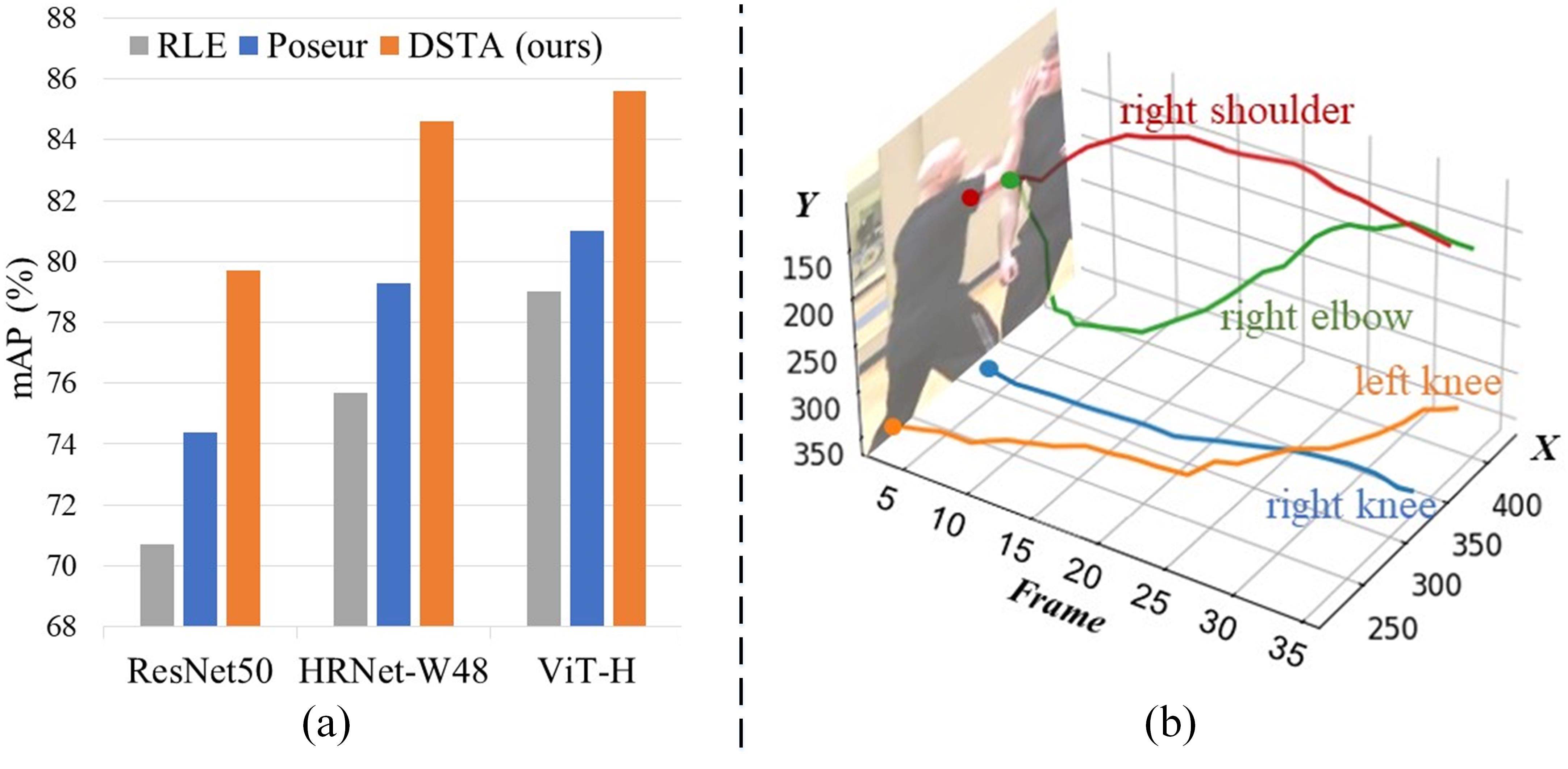}
    \caption{(a) Compared to our proposed video-based regression method, previous image-based regression methods of RLE~\cite{RLE_ICCV2021} and Poseur~\cite{Poseur_ECCV2022} have a substantial performance decline when processing video input,\textit{ e.g.}, the dataset of PoseTrack2017~\cite{PoseTrack2017_CVPR2017}. (b) Despite the intrinsic spatial correlations among human body joints, each joint exhibits independent motion trajectories temporally. }
    \label{fig:demo}
    \vspace{-5mm}
\end{figure}

Existing methods of human pose estimation can be divided into two categories: heatmap-based~\cite{Vitpose_NIPS2022,HRNet_CVPR2019,HRFormer_NIPS2021,CID_cvpr2022,FirstHeatmap_NIPS2014,SimplePose_ECCV2018,TokenPose_CVPR2023,DCPose_CVPR2021}, and regression-based~\cite{FirstRegression_CVPR2014,RLE_ICCV2021,Poseur_ECCV2022,Distilpose_CVPR2023,PRTR_CVPR2021}. Heatmap-based methods generate a likelihood
heatmap for each joint, whereas regression-based methods directly map the input to the output joint coordinates.
Owing to their superior performance, heatmap-based methods dominate in the field of human pose estimation, particularly among video-based approaches~\cite{PoseWarper_NIPS2019,DCPose_CVPR2021,OTPose_SMC2022,TDMI_CVPR2023}. The high computation and storage requirements of heatmap-based methods, however, make them expensive in 3D contexts (temporal), which restricts their versatility and real-time deployment in video applications, especially on edge devices. On the other hand, regression-based methods are more flexible and efficient. According to~\cite{RLE_ICCV2021}, while a standard heatmap head (3 deconv layers) costs $1.4\times$ FLOPs of the ResNet-50 backbone, the regression head only costs 1/20000 FLOPs of the same backbone. Moreover, recent regression-based approaches~\cite{RLE_ICCV2021,Poseur_ECCV2022} have demonstrated outstanding performance that is on par with heatmap-based methods.
Unfortunately, these regression-based approaches are all built for static images and neglect the temporal dependency between video frames, leading to a marked decline in performance when handling video input, as shown in Fig.~\ref{fig:demo}(a).

In this work, we explore the video-based human pose regression to facilitate multi-person pose estimation in video sequences. Regression-based approaches primarily focus on regressing the coordinates of pose joints, often overlooking the rich structural information inherent in the pose~\cite{CompositionalPose_ICCV2017}. As demonstrated in~~\cite{Poseur_ECCV2022}, the self-attention module employed in the Transformer architecture~\cite{Attention_NIPS2017} can be used across the pose joints to naturally capture their spatial dependency. A simple and direct extension is to use the self-attention module across all the joints from consecutive video frames to capture both the structural information of the pose and its temporal dependency in video sequences. 
However, as illustrated in Fig.~\ref{fig:demo}(b), while there's an inherent spatial correlation between adjacent joints of the human pose, the temporal trajectory of each joint tends to be rather independent. This implies that the spatial structure of the pose and its temporal dynamics across video frames cannot be conflated and must be captured separately.

To this end, we propose a novel and effective video-based human pose regression method, named Decoupled Space-Time Aggregation (DSTA), that
models the spatial structure between adjacent joints and the temporal dynamic of each individual joint separately, thereby avoiding the conflation of spatiotemporal information. Rather than using the output feature maps of a CNN backbone to regress the joints' coordinates as in existing regression models~\cite{RLE_ICCV2021,Poseur_ECCV2022}, DSTA converts the backbone's output into a sequence of tokens, with each token uniquely representing a joint.
Intuitively, each token embodies the feature embedding of its corresponding joint; therefore, it is natural to use them to model the spatiotemporal dependencies of pose joints. Specifically, DSTA first establishes the feature token for each joint via Joint-centric Feature Decoder (JFD)
module, which are hence used to capture the spatiotemporal relations of pose joints in the Space-Time Decoupling (STD) module. To efficiently and flexibly model the spatial dependency between adjacent joints and the temporal dependency of each joint itself, we introduce a joint-wise local-awareness attention mechanism to ensure each joint only attends to those joints that are structurally or temporally relevant. The aggregated spatial and temporal information is utilized to determine the coordinates of the joints. During training, the JFD and STD modules are optimized simultaneously, with the entire model undergoing end-to-end training.

To the best of our knowledge, this is an original effort on regression-based framework for multi-person pose estimation in video sequences.
We evaluate our method through the widely-utilized 
video-based benchmarks for human pose estimation: PoseTrack datasets~\cite{PoseTrack2017_CVPR2017,PoseTrack2018_CVPR2018,PoseTrack21_CVPR2022}.
With a simple yet effective architecture,
DSTA achieves a notable improvement of \textbf{8.9}
mAP over previous regression-based methods tailored for static images and obtains superior performance to the heatmap-based methods for video sequences. Moreover, it offers greater efficiency of computation and storage than heatmap-based multi-frame human pose estimation methods, making it more suitable for real-time video applications and easier to deploy, particularly on edge devices.
For instance, utilizing the HRNet-W48 backbone, our regression-based DSTA achieves \textbf{83.4} mAP on the PoseTrack2017~\cite{PoseTrack2017_CVPR2017} dataset with a head computation of merely \textbf{0.02} GFLOPs, while heatmap-based DCPose~\cite{DCPose_CVPR2021} attains 82.8 mAP on the same dataset with a significantly higher head computation of 11.0 GFLOPs.




Our main contributions can be summarized as follows: 
\begin{itemize}
    \item We propose \textbf{DSTA}, a novel and effective video-based human pose regression framework. The proposed method efficiently and flexibly models the spatiotemporal dependencies of pose joints in the video sequences.
    \item  Our method is the first regression-based method for multi-frame human pose estimation. Compared to heatmap-based methods, our method is efficient and flexible, opening up new possibilities for real-time video applications.
    \item We demonstrate the effectiveness of our approach with extensive experiments. Our method not only delivers a marked improvement over prior regression-based methods designed for static images, but also achieves performance superior to the heatmap-based methods.
\end{itemize}


\section{Related Work}
\label{sec:related_work}

\textbf{Heatmap-based Human Pose Estimation.}
Since the introduction of likelihood heatmaps to represent human joint positions~\cite{FirstHeatmap_NIPS2014}, heatmap-based methods have become predominant in the field of 2D human pose estimation~\cite{HRNet_CVPR2019,HRFormer_NIPS2021,Transpose_ICCV2021,UDPPose_cvpr2020,Vitpose_NIPS2022}, owing to their superior performance. To perform multi-person human pose estiamtion, the top-down approaches initially identify person bounding boxes and subsequently conduct single-person pose estimation within the cropped regions~\cite{maskrcnn_ICCV2017,HRNet_CVPR2019,Vitpose_NIPS2022}. Conversely, bottom-up methods commence by detecting identity-free keypoints for all individuals and then cluster these keypoints into distinct persons~\cite{pifpaf_CVPR2019,hrhrnet_cvpr2020,SWAHR_CVPR2021,CenterAttention_iccv2021}. Recently, the heatmaps or CNN features from adjacent frames have been utilized to extract the temporal dependencies of human poses, thereby enhancing the performance of multi-person human pose estimation in video sequences~\cite{PoseWarper_NIPS2019,DCPose_CVPR2021,FAMIPose_CVPR2022}. Despite its effectiveness, the heatmap representation inherently suffers from several drawbacks, such as quantization errors and the high computational and storage demands associated with maintaining high-resolution heatmaps.

\noindent \textbf{Regression-based Human Pose Estimation.} 
Regression-based methods forgo the use of intermediate heatmaps, opting instead to map the input directly to the output joint coordinates~\cite{FirstRegression_CVPR2014}. This approach is flexible and efficient for a wide range of human pose estimation tasks and real-time applications, especially on edge devices. Despite their efficiency, regression-based methods have traditionally lagged behind heatmap-based methods in accuracy within the realm of human pose estimation, leading to less focus on their development~\cite{FirstRegression_CVPR2014,PRTR_CVPR2021,SPM_ICCV2019,PointSetAnchor_ECCV2020}. Recently, advancements such as RLE~\cite{RLE_ICCV2021} and Poseur~\cite{Poseur_ECCV2022} have significantly propelled regression-based approaches, elevating their performance to a level comparable with heatmap-based methods. However, these regression-based methods are designed exclusively for static images. When these image-based methods are directly applied to video sequences, they tend to yield suboptimal predictions due to their inability to capture temporal dependencies between frames. As a result, such models struggle with challenges inherent to video inputs, such as motion blur, defocusing, and pose occlusions, which are common in dynamic scenes. 

In this work, we present for the first time a regression-based approach for multi-person human pose estimation in video sequences, outperforming or is on par with state-of-the-art heatmap-based methods for video sequences.

\section{Method}
\label{sec:method}

\subsection{Overview}
\label{sec:overview}

Given a  video frame $\mathcal{I}(t)$ at time $t$ containing multiple persons, we are interested in estimating locations of pose joints for each person. To enhance pose estimation for the  frame $\mathcal{I}(t)$, we 
leverage the temporal dynamics from a consecutive frame sequence $\mathbf{\mathcal{S}}=\langle\mathcal{I}({t-T}),\dots,\mathcal{I}(t),\dots,\mathcal{I}({t+T})\rangle$, where $T$ is a predefined temporal span. Our method follows the top-down paradigm. 
Initially, we use an human detector to identify individual persons in the frame $\mathcal{I}(t)$. Subsequently, each detected bounding box is expanded by 25\% to extract the same individual across the frame sequence $\mathbf{\mathcal{S}}$, resulting in a cropped video clip $\mathbf{\mathcal{S}}_i=\langle\mathcal{I}_i({t-T}),\dots,\mathcal{I}_i(t),\dots,\mathcal{I}_i({t+T})\rangle$ for every individual $i$. The goal of estimating human pose for individual $i$ within the specified video frame $\mathcal{I}(t)$ can then be denoted as
\begin{equation*}
      \{ \textbf{x}_i^j(t)  \}_{j=1}^n  = \text{HPE}({\mathcal{S}}_i),
\end{equation*}
where $\text{HPE}(\cdot)$ denotes the human pose estimation module, $\textbf{x}_i^j(t)$ is the $j$-th pose joint for the individual $i$ in video frame $\mathcal{I}(t)$, and $n$ represents the number of joints for each person, \textit{e.g.}, $n=15$ for PoseTrack datasets~\cite{PoseTrack2017_CVPR2017,PoseTrack2018_CVPR2018,PoseTrack21_CVPR2022}. 
For simplicity in the following description of our algorithm, unless otherwise specified, we will refer to a specific individual $i$.

\begin{figure*}
    \centering
    \includegraphics[width=1.0\linewidth]{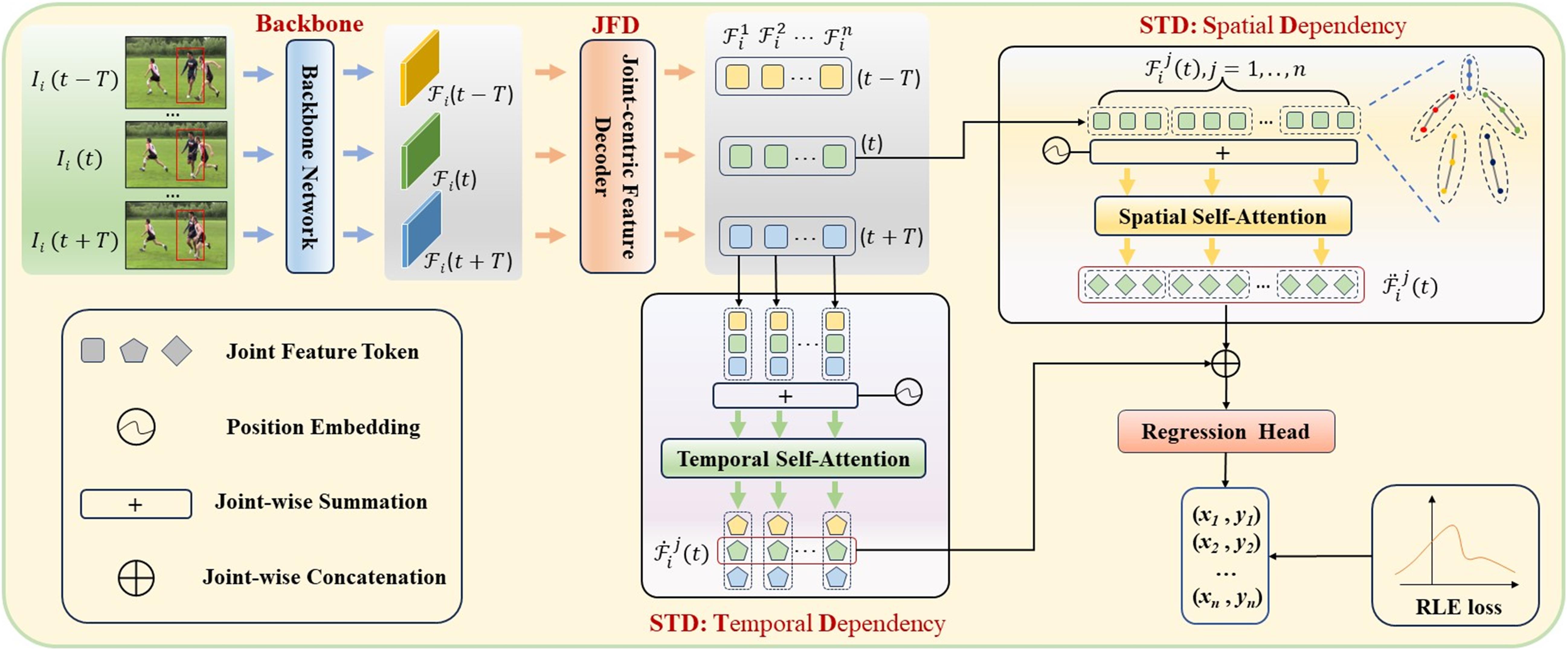}
    \caption{The pipeline of the proposed Decoupled Space-Time Aggregation (DSTA). The goal is to detect the human pose of the key frame $\mathcal{I}_i(t)$. Given a video sequence $\langle\mathcal{I}_i({t-T}),\dots,\mathcal{I}_i(t),\dots,\mathcal{I}_i({t+T})\rangle$, DSTA uses a backbone network to extract their feature maps. From these maps, Joint-centric Feature Decoder (JFD) extracts feature tokens to individually represent each joint. Space-Time Decoupling (STD) then models the temporal dynamic dependencies and spatial structural dependencies of joints separately, producing aggregated space-time features for the current key frame. Each of these aggregated features is  utilized to regress the coordinates of the corresponding joint.}
    \label{fig:pipeline}
    \vspace{-2mm}
\end{figure*}

We adopt the regression-based method to implement the human pose estimation module $\text{HPE}(\cdot)$. Compared to the heatmap-based method, the regression-based method offers several advantages: i) It eliminates the need for high-resolution heatmaps, resulting in reduced computation and storage demands. This makes it more apt for real-time video applications and facilitates deployment, especially on edge devices. ii) It provides continuous outputs, avoiding the quantization issues inherent in heatmap methods. In the regression-based pose estimation module $\text{HPE}(\cdot)$, the global feature maps extracted by a CNN backbone are fed into a regression module, which then directly produces the coordinates of the joints, \textit{i.e.},  
\begin{equation}
    \{ \textbf{x}_i^j(t)  \}_{j=1}^n = \text{REG}(\hat{\mathcal{S}}_i),
\end{equation}
where $\text{REG}(\cdot)$ denotes the regression module and $\hat{\mathcal{S}}_i=\langle\mathcal{F}_i({t-T}),\dots,\mathcal{F}_i(t),\dots,\mathcal{F}_i({t+T})\rangle$. Here, $\mathcal{F}_i(t')$ with $t'\in[t-T,t+T]$ is the global feature maps extracted by the CNN backbone from the cropped image $\mathcal{I}_i(t')$. Note, when regressing the pose joints at the current frame time $t$, we aim to utilize the temporal feature information across the video clip $\mathbf{\mathcal{S}}_i$, rather than solely relying on the feature information from the current frame $\mathcal{I}_i(t)$.

Prior work, such as Poseur~\cite{Poseur_ECCV2022}, has demonstrated that the spatial dependencies among pose joints can be naturally captured by applying the self-attention mechanism~\cite{Attention_NIPS2017} over them. It follows that we can also employ the self-attention mechanism on the global pose features in the temporal sequence $\hat{\mathcal{S}}_i$ to discern the temporal dependency of individual's pose over the time interval $[t-T, t+T]$.
As depicted in Fig.~\ref{fig:demo}(b), each pose joint exhibits a relatively independent temporal trajectory. Hence, it's more appropriate to model the temporal dependency at the joint level rather than for the entire pose. To this end, extra efforts are required to convert each global feature  $\mathcal{F}_i(t')$ into a set of joint-aware feature embeddings. This procedure is finished in
the Joint-centric Feature Decoder (JFD) , which
can be denoted as
\begin{equation}
    \{  \mathcal{F}_i^j(t')  \}_{j=1}^n = \text{JFD}(\mathcal{F}_i(t')), \quad t'\in[t-T, t+T],
\end{equation}
where $\mathcal{F}_i^j(t')$, termed a feature token, represents the feature embedding of the $j$-th joint of the pose at the video frame of time $t'$. 
Let's represent the feature tokens for each joint of the pose over the time span $[t-T, t+T]$ as $\Tilde{\mathcal{S}}_i$. That is, $\Tilde{\mathcal{S}}_i =  \langle \{  \mathcal{F}_i^j({t-T})  \}_{j=1}^n,\dots,\{  \mathcal{F}_i^j({t})  \}_{j=1}^n,\dots,\{  \mathcal{F}_i^j({t+T})  \}_{j=1}^n  \rangle.$
We subsequently utilize the feature tokens in $\Tilde{\mathcal{S}}_i$ to model the spatiotemporal dependencies of pose joints via Space-Time Decoupling (STD),
\begin{equation}
    \{  \textbf{f}_i^j(t)  \}_{j=1}^n = \text{STD}(\Tilde{\mathcal{S}}_i),
\end{equation}
where $\textbf{f}_i^j(t)$ is the aggregated space-time feature for the $j$-th joint of the pose at the current video frame $t$, which is then fed to a joint-wise fully connected feed-forward network to produce the coordinates of the joint. 

Our proposed Decoupled Space-Time Aggregation Network (DSTA) is composed of three primary modules: the backbone, $\text{JFD}(\cdot)$, and $\text{STD}(\cdot)$. We learn DSTA by training the backbone, $\text{JFD}$,  $\text{STD}$ modules in an end-to-end manner.
The architecture and workflow of DSTA are depicted in Fig.~\ref{fig:pipeline}. Subsequent sections delve into the specifics of the JFD, STD, and the computation of the loss function.

\subsection{Joint-centric Feature Decoder}
\label{sec:JFD}

As shown in Fig.~\ref{fig:pipeline}, the purpose of JFD is to extract the feature embedding for each joint from the given global feature maps $\mathcal{F}_i(t')$ with $t'\in[t-T,t+T]$. As suggested by~\cite{Poseur_ECCV2022}, one potential approach to construct the joint embedding is as follows: initially, a traditional regression method such as~\cite{RLE_ICCV2021} is utilized to regress the joint coordinates from $\mathcal{F}_i(t')$. For each joint, its $x$-$y$ coordinates are converted into position embedding using sine-cosine position encoding~\cite{Attention_NIPS2017}. Concurrently, a learnable class embedding is designated for every joint type. The final feature embedding for each joint is derived by summing its position embedding with the respective class embedding. However, this approach loses crucial contextual information of the joints within the pose that is learned in the global feature maps $\mathcal{F}_i(t')$. Though we can augment each joint with relevant contextual feature from the global feature maps, such as the approach in~\cite{Poseur_ECCV2022} which uses the joint's embedding as a query and applies a multi-scale deformable attention module to sample features for each joint from the feature maps, this method incurs significant computational costs.


We employ a straightforward yet efficient approach to construct the joint embeddings from the provided global feature maps $\mathcal{F}_i(t')$.
Given the global feature maps produced by the backbone, previous heatmap-based methods convolve these maps via convolution layers to generate a heatmap feature for each joint~\cite{PoseWarper_NIPS2019,DCPose_CVPR2021}. 
We follow this strategy, deriving the feature embedding for each joint from $\mathcal{F}_i(t')$ through a convolution layer or a fully connected layer (FC). 
In our setup, the ResNet backbones (like ResNet50 or ResNet152) are followed by a global average pooling layer and a FC layer. The FC layer comprises $2048 \times K$ neurons. Here, 2048 represents the dimensionality of $\mathcal{F}_i(t')$ after undergoing global average pooling and flattening. Meanwhile, $K$ is calculated as $n \times 32$, where $n$ indicates the number of pose joints, and 32 signifies the dimension of the joint embedding.
The output of the FC layer is the feature embedding for each joint, where the output is evenly divided into 
$n$ parts, denoted as $\{  \mathcal{F}_i^j(t')  \}_{j=1}^n$, with each part representing a feature embedding for a joint. Further implementation details regarding more backbones (\textit{e.g.}, HRNet backbone) can be found in the supplementary material.

\subsection{Space-Time Decoupling}
\label{sec:STD}

STD is designated to model the spatial and temporal dependencies between joints based on their embeddings over the time span $[t-T, t+T]$,  \textit{i.e.}, all feature tokens in $\Tilde{\mathcal{S}}_i =  \langle \{  \mathcal{F}_i^j({t-T})  \}_{j=1}^n,\dots,\{  \mathcal{F}_i^j({t})  \}_{j=1}^n,\dots,\{  \mathcal{F}_i^j({t+T})  \}_{j=1}^n  \rangle$. 
In numerous applications~\cite{Bert_NAACL2019,Poseur_ECCV2022}, the self-attention mechanism's proficiency in capturing long-distance dependencies within sequences has been thoroughly demonstrated~\cite{Attention_NIPS2017}. Thus, a direct approach to  capturing the spatio-temporal dependencies between joints is to apply the self-attention module to the sequence of feature tokens in $\Tilde{\mathcal{S}}_i$,
\begin{equation}
\label{eqn:simple_dependency}
 \{  \mathcal{\hat{F}}_i^j({t})  \}_{j=1}^n =   \text{S-ATT}(\Tilde{\mathcal{S}}_i),
\end{equation}
where $\text{S-ATT}(\cdot)$ denotes the self-attention module~\cite{Attention_NIPS2017}, and $\mathcal{\hat{F}}_i^j({t})$, which encodes the spatial and temporal information learned by the S-ATT module, represents the updated feature token for the $j$-th joint at the current video frame $t$. 
In our implementation, the S-ATT module adheres to the conventional Transformer architecture~\cite{Attention_NIPS2017}. In our setup, $4$ identical layers are stacked sequentially. 
Each layer comprises two sub-layers: the first one employs a multi-head self-attention mechanism, and the second one utilizes a simple, token-wise, fully connected feed-forward network. The input feature tokens  pass through these modules in sequence, each producing an updated version that serves as the input for the subsequent layer. Additionally, each of the initial input feature tokens is equipped with a learnable position embedding, and their  sum forms the final input. 

\subsubsection{Decoupled Space-Time Aggregation}
\label{sec:dsta}
However, as illustrated in Fig.~\ref{fig:demo}(b), despite the inherent spatial correlation among adjacent joints of the human pose, the temporal trajectory of each individual joint tends to be rather independent. So, as shown in Fig.~\ref{fig:pipeline}, our proposed DSTA models the temporal dynamic dependencies and spatial structure dependencies separately, instead of modeling spatial and temporal dependencies together as in Eq.~\ref{eqn:simple_dependency}. This approach allows for a more nuanced capture of the unique dependency characteristics that joints exhibit separately in both the temporal and spatial dimensions.
Then, by fusing the captured spatial and temporal information, an aggregated spatio-temporal feature for each joint of current frame $t$, \textit{i.e.}, $\textbf{f}_i^j({t})$, is derived:
\begin{equation}
\label{eqn:decouple_dependency}
 \{ \textbf{f}_i^j({t})  \}_{j=1}^n =   \text{SD}(\Tilde{\mathcal{S}}_i) \bigoplus \text{TD}(\Tilde{\mathcal{S}}_i),
\end{equation}
where $\bigoplus$ denotes the concatenation operation, which is individually applied to each pair of corresponding updated feature tokens associated with each joint. 
By utilizing the local-awareness attention introduced below (Sec.~\ref{sec:local_attention}), the $\text{SD}(\cdot)$ module learns the spatial dependencies between adjacent joints and correspondingly generates an updated feature token for each joint in the current frame.  Concurrently,  the $\text{TD}(\cdot)$ module discerns the temporal dependencies of each joint, resulting in another updated feature token for each joint in the current frame.
Subsequently, the aggregated features of joints  
$\{ \textbf{f}_i^j({t})  \}_{j=1}^n$ are fed into a joint-wise fully connected feed-forward network, 
producing the coordinates of the joints $\{ \textbf{x}_i^j(t)  \}_{j=1}^n$:
\begin{equation}
    \{ \textbf{f}_i^j({t})  \}_{j=1}^n \xrightarrow[\text{feed-forward network}]{\text{joint-wise fully connected}} \{ \textbf{x}_i^j(t)  \}_{j=1}^n.
\end{equation}

\subsubsection{Local-awareness Attention}
\label{sec:local_attention}
From a temporal perspective, each joint is intimately connected only with its corresponding joints in preceding and succeeding frames, having no relevance with other joints. From a spatial perspective, the structure dependencies of joints are primarily manifested between adjacent joints within a single frame. Therefore, we introduce a joint-wise local-awareness attention mechanism, ensuring that each joint only attends to those that are structurally or temporally relevant. This local-awareness attention mechanism is elaborated upon, demonstrating its application in implementing the aforementioned $\text{SD}(\cdot)$ and $\text{TD}(\cdot)$ modules.

In the $\text{TD}$ module, we capture the temporal dynamic dependency for each joint $j$ at the current frame $t$. To this end, our proposed local-awareness attention selectively applies the self-attention module $\text{S-ATT}$ in Eq.~\ref{eqn:simple_dependency} across the corresponding joints over the time span $[t-T, t+T]$, 
\begin{equation}
\label{eqn:time}
  \mathcal{\dot{F}}_i^j({t}) = \text{S-ATT}( \Tilde{\mathcal{S}}_i^j ), \quad j=1,2,\dots,n,
\end{equation}
where $\Tilde{\mathcal{S}}_i^j=\langle \mathcal{F}_i^j({t-T}),\dots, \mathcal{F}_i^j({t}),\dots,\mathcal{F}_i^j({t+T}) \rangle$, and $\mathcal{\dot{F}}_i^j({t})$ denotes the updated feature token for the $j$-th joint at the current video frame $t$, encoding the temporal dependency information of this joint embedded within the sequence $\Tilde{\mathcal{S}}_i^j$. 
Since the sequence $\Tilde{\mathcal{S}}_i^j$ only includes the feature tokens of joint $j$ over the time span $[t-T, t+T]$, the temporal dependency encoded in $\mathcal{\dot{F}}_i^j({t})$ is solely related to the joint itself, without any relevance to other joints.

In the $\text{SD}$ module, we capture the spatial structure dependency among joints within the current frame $t$. A straightforward way is to directly apply the self-attention module $\text{S-ATT}$ from Eq.~\ref{eqn:simple_dependency} to all joints in the current frame. 
To allow each joint to focus more closely on the adjacent joints that are intimately associated with it in structure, we divide the joints into $K$ groups according to the semantic structure of the human pose, as shown in the top right of Fig.~\ref{fig:pipeline}. Our proposed local-awareness attention conducts the self-attention module $\text{S-ATT}$ separately for each group,
\begin{equation}
\label{eqn:space}
 \{ \mathcal{\ddot{F}}_i^{j}({t}) \}_{j \in G(k)}  = \text{S-ATT}( \langle  {\mathcal{F}}(t)_i^{j}  \rangle_{j \in G(k)} ),\quad k=1,\dots,K,
\end{equation}
where $G(k)$ represents the set of joint indices in group $k$, and $\mathcal{\ddot{F}}_i^{j}({t})$ denotes the updated feature token for the $j$-th joint at the current video frame $t$, encapsulating the spatial structure dependencies of this joint within the pose. 

Through the modules $\text{TD}$ and $\text{SD}$ , we have captured the spatial and temporal contexts for each joint in the current frame, obtaining the corresponding updated feature tokens, $\mathcal{\dot{F}}_i^{j}({t})$ and $\mathcal{\ddot{F}}_i^{j}({t})$.  Consequently, the spatio-temporal aggregated feature $\textbf{f}_i^j({t})$ for each joint $j$ at the current frame $t$, as per Eq.~\ref{eqn:decouple_dependency}, can be explicitly computed as follows:
\begin{equation}
\label{eqn:feature_aggregation}
  \textbf{f}_i^j({t})  =   \mathcal{\dot{F}}_i^{j}({t}) \oplus \mathcal{\ddot{F}}_i^{j}({t}),\quad j=1,2,\dots,n,
\end{equation}
where $\oplus$ denotes the concatenation operation. 

\textbf{Discussion: } Compared with the \textit{global} attention method as defined in Eq.~\ref{eqn:simple_dependency}, our proposed \textit{ local-awareness} attention ensures that each joint only attends to those that are structurally or temporally relevant. This approach not only avoids the undesired conflation of spatiotemporal dimensions but also reduces computational overhead. 
For example, the computational cost of the \mbox{S-ATT} module is mainly determined by the multi-head self-attention mechanism~\cite{Attention_NIPS2017}, where the computational complexity is proportionate to the square of the quantity of feature tokens. 
Consequently, the computational complexity of the global attention method delineated in Eq.~\ref{eqn:simple_dependency} is approximately \( O((n \times (2T+1))^2) = O(4n^2T^2+4n^2T+n^2) \). In contrast, the total computational complexity of our local-awareness attention methods, corresponding to Eqs.~\ref{eqn:time} and \ref{eqn:space}, is approximately \( O( n \times (2T+1)^2 + K \times (\frac{n}{K})^2 ) = O(4nT^2+4nT+n+\frac{n^2}{K}) \). In the experiment, the value of $T$ is quite small, for instance $T=1$, thus our local-awareness attention method reduces the time complexity from \(  O(9n^2) \) to \( O(\frac{n^2}{K}+9n) \) with $K=5$ in our implementation, thereby achieving a speedup close to 45 times.

\subsection{Loss Computation}
\label{sec:loss}
During training, the entire model undergoes end-to-end optimization, aiming to minimize the discrepancy between the coordinates of the predicted joints and the ground truth joints in the current frame $t$. To boost the regression performance, we employ the residual log-likelihood estimation loss (RLE) as proposed in~\cite{RLE_ICCV2021}, in lieu of the conventional regression loss ($l_1$ or $l_2$). We extend the RLE loss, originally designed for image-based pose regression, to the context of video-based pose regression.
Given an input cropped video clip, $\mathcal{S}_i$, for individual $i$, we calculate a  distribution, $P_{\theta,\phi}( \{ \textbf{x}_i^j(t) \}_{j=1}^n | \mathcal{S}_i )$, which reflects the likelihood  that the ground truth at the current frame $t$ appears at the predicted locations $\{ \textbf{x}_i^j(t)  \}_{j=1}^n$. Here, $\theta$ represents the parameters of our model, and $\phi$ represents the parameters of a flow model. The flow model is not required to operate during inference, thereby introducing no additional overhead at test time. The learning process involves the simultaneous optimizations of the model parameters $\theta$ and $\phi$, aiming to maximize the probability of observing the ground truth $\mu_g$. This is achieved by defining the RLE loss as follows:
\begin{equation}
    \mathcal{L}_{rle}= -P_{\theta,\phi}( \{ \textbf{x}_i^j(t) \}_{j=1}^n | \mathcal{S}_i) \Big|_{ \{ \textbf{x}_i^j(t) \}_{j=1}^n = \mu_g }.
\end{equation}
For a more detailed discussion and further information about the  RLE loss, we refer readers to~\cite{RLE_ICCV2021}.

\section{Experiments}
\label{sec:experiments}

\subsection{Experimental Settings}
We have conducted evaluations on three widely-utilized video-based benchmarks for human pose estimation: PoseTrack2017~\cite{PoseTrack2017_CVPR2017}, PoseTrack2018~\cite{PoseTrack2018_CVPR2018}, and PoseTrack21~\cite{PoseTrack21_CVPR2022}. These datasets contain video sequences of complex scenarios involving rapid movements of highly occluded individuals in crowded environments. 
To assess the performance of our models, we utilize the Average Precision (AP) metric~\cite{HRNet_CVPR2019,PoseWarper_NIPS2019,DCPose_CVPR2021,RLE_ICCV2021}. The AP is calculated for each joint, and the mean AP across all joints is denoted as mAP.
Our method is implemented using PyTorch. 
Unless otherwise specified, the input image size is $384 \times 288$ when using the HRNet-w48 backbone, while for other backbones, the input image size is $256 \times 192$.
We pretrained the backbones on the COCO dataset. For further implementation details, please refer to the supplementary materials provided.


\subsection{Main Results}

\begin{table}
    \centering
    \begin{tabular}{l|ccc}
       Method  &  ResNet-50 &  HRNet-W48   &  ViT-H \\
       \hline
       \multicolumn{4}{l}{\textit{image-based}}   \\
       RLE~\cite{RLE_ICCV2021}  &  70.7  &  75.7  &  79.0 \\
       Poseur~\cite{Poseur_ECCV2022}  & 74.4  & 79.3 &  81.0 \\
       \hline
       \multicolumn{4}{l}{\textit{video-based}} \\
       \textbf{DSTA (Ours)}  &  \textbf{79.7}  &  \textbf{84.6}  &  \textbf{85.6}  \\
    \end{tabular}
    \vspace{-2mm}
    \caption{\textbf{Comparison with image-based regression} (mAP) on PoseTrack2017 val. set.}
    \label{tab:regression_compare}
    \vspace{-5mm}
\end{table}

\subsubsection{Comparison with Image-based Regression}
To study the effectiveness of the proposed regression method on video input, we compare it with existing state-of-the-art image-based regression methods, namely RLE~\cite{RLE_ICCV2021} and Poseur~\cite{Poseur_ECCV2022}.
For thorough and fair comparisons, we utilized three distinct backbone networks—ResNet-50, HRNet-W48, and ViT-H—and ensured that each approach applied the identical pre-trained model to each backbone network. The experimental results on the PoseTrack2017 validation set are presented in Table~\ref{tab:regression_compare}. As shown, our proposed video-based method achieves significant performance improvements across all backbone networks when compared to image-based methods. For instance, our method outperforms RLE~\cite{RLE_ICCV2021} by a notable margin of \textbf{8.9} mAP (or \textbf{9.0} mAP) when utilizing the HRNet-W48 (or ResNet-50) backbone. This demonstrates the importance of incorporating temporal cues from neighboring frames.

\begin{figure}
    \centering
    \includegraphics[width=1.0\linewidth]{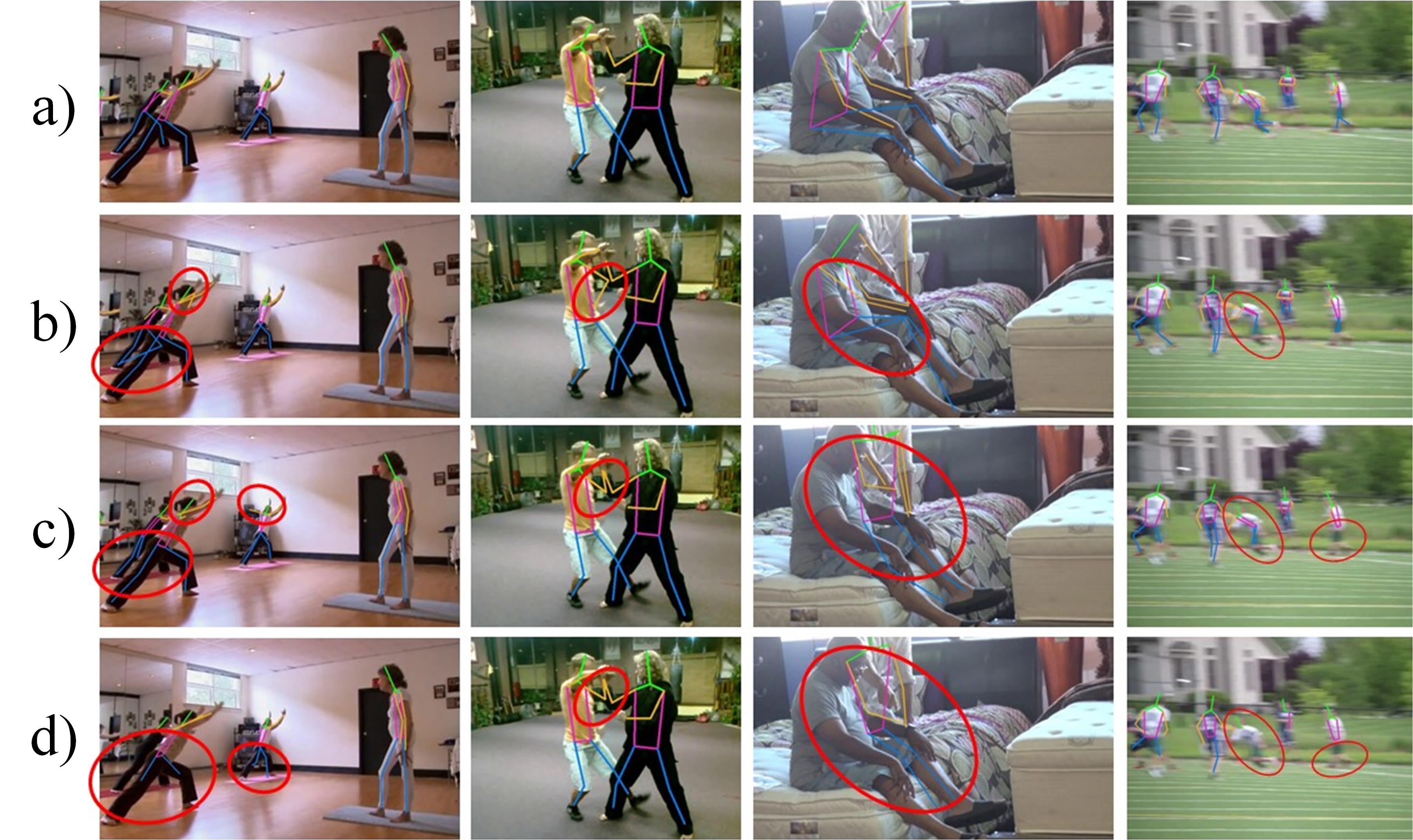}
    \caption{\textbf{Qualitative comparison} of a) our DSTA, b) DCPose~\cite{DCPose_CVPR2021}, c) Poseur~\cite{Poseur_ECCV2022}, and d) RLE~\cite{RLE_ICCV2021} on the PoseTrack datasets, featuring challenges such as occlusions, nearby-person interactions, and motion blur. Inaccurate predictions are marked
    with red solid circles.}
    \label{fig:qualitative_result}
    \vspace{-5mm}
\end{figure}

By leveraging temporal dependencies across consecutive frames, our video-based regression is better equipped to handle challenging situations such as occlusion or motion blur commonly encountered in video scenarios, as demonstrated in Fig.~\ref{fig:qualitative_result}. These experiments demonstrate the superior performance of our proposed video-based regression framework, which significantly outstrips the capabilities of prior image-based methods when handling the video input. 

\subsubsection{Comparison with State-of-the-art Methods}
Current state-of-the-art algorithms for video-based human pose estimation are predominantly based on heatmaps. We first conduct a performance comparison of our method with these heatmap-based approaches on the PoseTrack datasets. Subsequently, we compare the computational complexity between the heatmap-based methods and our proposed regression-based approach. Furthermore, we examine the varying impacts of input resolution on both the heatmap-based methods and our regression-based approach.

\textbf{Results on the PoseTrack Datasets.} 
Table~\ref{tab:sota_compare} presents the quantitative results of different approaches on PoseTrack2017 validation set.
Our method achieves comparable performance to the state-of-the-art methods.
For example, employing the HRNet-W48 backbone, our method attains an mAP of \textbf{84.6}, which surpasses the adopted backbone network HRNet-W48~\cite{HRNet_CVPR2019} by \textbf{7.3} points. When compared to video-based approaches utilizing the same backbone, our method outperforms DCPose~\cite{DCPose_CVPR2021} by \textbf{1.8} points while maintaining a performance level on par with the state-of-the-art FAMI-Pose~\cite{FAMIPose_CVPR2022}.
Our approach is flexible and can be easily integrated into various backbone networks. When using the ViT-H backbone, our method further pushes the performance boundary and achieves an mAP of \textbf{85.6}. The performance enhancement for the relatively challenging joints is truly encouraging: an mAP of \textbf{82.6} ($\uparrow$ \textbf{2.6}) for wrists and an mAP of \textbf{77.8} ($\uparrow$ \textbf{0.8}) for ankles. 
It is worth noting that methods utilizing temporal information, such as PoseWarper~\cite{PoseWarper_NIPS2019}, DCPose~\cite{DCPose_CVPR2021}, DetTrack~\cite{DetTrack_cvpr2020}, FAMI-Pose~\cite{FAMIPose_CVPR2022}, and our DSTA, consistently outperform those relying solely on a single key frame, such as HRNet~\cite{HRNet_CVPR2019}. This reaffirms the importance of incorporating temporal cues from adjacent frames.  Qualitative results are shown in Fig.~\ref{fig:qualitative_result}.

\begin{table}
\footnotesize
    \centering
    \setlength{\tabcolsep}{0.5mm}{
    \begin{tabular}{l|c|ccccccc|c}
       Method  & Bkbone & Head	 & Should. & Elbow	& Wrist & Hip & Knee & Ankle & \textbf{Mean} \\
       \hline
       \multicolumn{10}{l}{\textit{heatmap-based}}   \\
       \fontsize{6}{1}\selectfont PoseTrack~\cite{PoseTrack_cvpr2018} & \fontsize{6}{1}\selectfont ResNet-101 &  67.5	& 70.2	& 62.0	& 51.7 & 60.7 &	58.7 & 49.8 & 60.6 \\
       \fontsize{6}{1}\selectfont PoseFlow~\cite{PoseFlow_arXiv2018} & \fontsize{6}{1}\selectfont ResNet-152 & 66.7  & 73.3 & 	68.3 & 	61.1 & 	67.5 & 	67.0 & 	61.3 & 	66.5 \\
       \fontsize{6}{1}\selectfont FastPose~\cite{fastpose_arXiv2019} & \fontsize{6}{1}\selectfont ResNet-101 & 	80.0 & 	80.3 & 	69.5 & 	59.1 & 	71.4 & 	67.5 & 	59.4 & 	70.3 \\
       \fontsize{6}{1}\selectfont SimBase.~\cite{SimplePose_ECCV2018} & \fontsize{6}{1}\selectfont ResNet-152 & 	81.7 &	83.4 &	80.0 &	72.4 &	75.3 &	74.8 &	67.1 &	76.7 \\
       \fontsize{6}{1}\selectfont STEmbed.~\cite{STE_CVPR2019} & \fontsize{6}{1}\selectfont ResNet-152 &	83.8 &	81.6 &	77.1 &	70.0 &	77.4 &	74.5 &	70.8 &	77.0 \\
       \fontsize{6}{1}\selectfont HRNet~\cite{HRNet_CVPR2019} & \fontsize{6}{1}\selectfont HRNet-W48 &	82.1 &	83.6 &	80.4 &	73.3 &	75.5 &	75.3 &	68.5 & 77.3 \\
       \fontsize{6}{1}\selectfont MDPN~\cite{MDPN_ECCV-W2018} & \fontsize{6}{1}\selectfont ResNet-152 &	85.2 &	88.5 &	83.9 &	77.5 &	79.0 &	77.0 &	71.4 & 80.7 \\
       \fontsize{6}{1}\selectfont Dyn.-GNN~\cite{DynGNN_CVPR2021} & \fontsize{6}{1}\selectfont HRNet-W48 &	88.4 &	88.4 &	82.0 &	74.5 &	79.1 &	78.3 &	73.1 &	81.1 \\
       \fontsize{6}{1}\selectfont PoseWarp.~\cite{PoseWarper_NIPS2019} & \fontsize{6}{1}\selectfont HRNet-W48 & 81.4 &	88.3 &	83.9 &	78.0 &	82.4 &	80.5 &	73.6 &	81.2 \\
       \fontsize{6}{1}\selectfont DCPose~\cite{DCPose_CVPR2021} & \fontsize{6}{1}\selectfont HRNet-W48 &	88.0 &	88.7 &	84.1 &	78.4 &	83.0 &	81.4 &	74.2 &	82.8 \\
       \fontsize{6}{1}\selectfont DetTrack~\cite{DetTrack_cvpr2020} & \fontsize{6}{1}\selectfont HRNet-W48 &	89.4 &	89.7 &	85.5 &	79.5 &	82.4 &	80.8 &	76.4 &	83.8 \\
      \fontsize{6}{1}\selectfont FAMIPose~\cite{FAMIPose_CVPR2022} & \fontsize{6}{1}\selectfont HRNet-W48 &	89.6 &	90.1 &	\textbf{86.3} &	\textbf{80.0} &	84.6 &	\textbf{83.4} &	\textbf{77.0} &	\textbf{84.8} \\
       \hline
       \multicolumn{10}{l}{\textit{regression-based}} \\
       \fontsize{6}{1}\selectfont \textbf{DSTA (Ours)} & \fontsize{6}{1}\selectfont ResNet-152  &88.3 &	88.1 &	83.3 &	76.0 &	82.5 &	81.1 &	70.0 &	81.8 \\
      \fontsize{6}{1}\selectfont \textbf{DSTA (Ours)} &  \fontsize{6}{1}\selectfont HRNet-W48  &  \textbf{89.8} &	\textbf{90.8} &	\textbf{86.2} &	79.3 &	\textbf{85.2} &	82.2 & 75.9 & \textbf{84.6} \\
      
       \cellcolor{light-gray}\fontsize{6}{1}\selectfont \textbf{DSTA (Ours)} & \cellcolor{light-gray}\fontsize{6}{1}\selectfont ViT-H  &	\cellcolor{light-gray}89.3 &	\cellcolor{light-gray}90.6 &	\cellcolor{light-gray}87.3 &	\cellcolor{light-gray}82.6 &	\cellcolor{light-gray}84.5 &	\cellcolor{light-gray}85.1 & \cellcolor{light-gray}77.8 &	\cellcolor{light-gray}85.6 \\
    \end{tabular}}
    \vspace{-2mm}
    \caption{\textbf{Comparison with the SOTA} on PoseTrack2017 val. set. Similar to FAMI-Pose~\cite{FAMIPose_CVPR2022}, our proposed DSTA sets the temporal span $T$ to 2, consisting of two preceding and two subsequent frames, totalling four auxiliary frames.}
    \label{tab:sota_compare}
\end{table}

We further evaluate our model on the PoseTrack2018 and PoseTrack21 datasets.
Due to space limitation, we present these results in the supplementary materials. 
Based on these results, it is evident that our approach either outperforms or is on par with the state-of-the-art heatmap-based methods. Using the HRNet-w48 backbone, we achieve \textbf{82.1} mAP and \textbf{82.0} mAP on these two datasets, respectively, while using the ViT-H backbone, we further improve performance by \textbf{1.3} and \textbf{1.5} points, respectively.

\begin{table}
\small
    \centering
    \setlength{\tabcolsep}{1mm}{
    \begin{tabular}{l|cccc}
       \multirow{2}{*}{Method}  &  \multirow{2}{*}{\#Params} &  GFLOPs   &  GFLOPs & \multirow{2}{*}{mAP} \\
       & & of Backbone & of Net. Head & \\
       \hline
       \multicolumn{5}{l}{\textit{heatmap-based}}   \\
       PoseWarper~\cite{PoseWarper_NIPS2019}  &  71.1M  &  35.5  &  156.7 & 81.0 \\
       DCPose~\cite{DCPose_CVPR2021}  & 65.2M   & 35.5  & 11.0  &  82.8 \\
       \hline
       \multicolumn{5}{l}{\textit{regression-based}} \\
       \textbf{DSTA (Ours)}  &  63.9M  &  35.5  & \textbf{0.02} & \textbf{83.4}  \\
    \end{tabular}}
    \vspace{-2mm}
    \caption{\textbf{Computation complexity} with HRNet-W48 backbone. \#Params includes the parameters of entire network.  All methods utilize the same two auxiliary frames as  in~\cite{DCPose_CVPR2021}.}
    \label{tab:complexity_compare}
    \vspace{-5mm}
\end{table}

\begin{table*}
\small
\parbox{.37\linewidth}{
    \centering
    \setlength{\tabcolsep}{1.5mm}{
    \begin{tabular}{c|c|cc|ccc}
       \multicolumn{2}{c|}{JFD}      &  w/o   & w/  & \checkmark  & \checkmark &  \checkmark  \\
       \hline
       \multirow{2}{*}{STD} & SD     &     &     &  \checkmark   &  & \checkmark  \\
        & TD                         &  &  &  & \checkmark & \checkmark  \\
        \hline
        \multicolumn{2}{c|}{mAP}     & 73.8 & 74.8 & 71.4 & 78.1 & \textbf{78.6} \\
    \end{tabular}}
    \caption{\textbf{Ablation of different modules} in DSTA.}
    \label{tab:impact_components}
}
\parbox{.25\linewidth}{
\centering
\setlength{\tabcolsep}{1mm}{
    \begin{tabular}{c|cc}
       JFD Method  &  MFLOPs &  mAP \\
       \hline
       \cite{Poseur_ECCV2022} (a) & 0.3	& 74.6   \\
       \cite{Poseur_ECCV2022} (b)  & 19.3	& 77.9 \\
       \textbf{Ours}  & 5.0 &	\textbf{78.6} \\
    \end{tabular}}
    \caption{\textbf{Different methods for constructing joint embeddings.}}
    \label{tab:jfd_choice}
}
\parbox{.38\linewidth}{
 \centering
    \setlength{\tabcolsep}{1mm}{
    \begin{tabular}{c|ccc}
       \#Auxiliary Frame  & ResNet-50 & HRNet-W48 & ViT-H  \\
       \hline
       1 \{-1\} & 78.0 & 82.6 & 82.6 \\
      2 \{-1, +1\} &	78.6 & 83.4	& 84.3 \\
      4 \{-2, -1, +1, +2\} & \textbf{79.7}  & \textbf{84.6} &	\textbf{85.6} \\
    \end{tabular}}
    \vspace{-3mm}
    \caption{\textbf{Different number of auxiliary frames}. `-' indicates previous frames while `+' indicates subsequent frames.}
    \label{tab:t_selection}
}
\vspace{-3mm}
\end{table*}

\begin{table}
\small
    \centering
    \setlength{\tabcolsep}{2mm}{
    \begin{tabular}{l|cccc}
    \hline
       \multirow{2}{*}{Method}  &  \multicolumn{4}{c}{Input Size}\\
       & 384×288  & 256×192 & 128×128 & 64×64 \\
       \hline
       \multicolumn{5}{l}{\textit{heatmap-based}}   \\
       DCPose~\cite{DCPose_CVPR2021}  & 82.8   & 81.2  & 71.7  &  35.1 \\
       \hline
       \multicolumn{5}{l}{\textit{regression-based}} \\
       \textbf{DSTA (Ours)}  &  \textbf{83.4} & \textbf{82.3} & \textbf{77.9} & \textbf{55.4}  \\
    \end{tabular}}
    \vspace{-2mm}
    \caption{\textbf{Performance with different input resolutions}. Note that, as in~\cite{DCPose_CVPR2021}, only two auxiliary frames are used in DSTA.}
    \label{tab:resolution_compare}
    \vspace{-5mm}
\end{table}

\textbf{Computation Complexity.}
We conduct experiments to assess computation complexity using the PoseTrack2017 validation set, and the results are presented in Table~\ref{tab:complexity_compare}.
To ensure a fair comparison, all methods utilize the identical HRNet-W48 backbone and adopt the identical two auxiliary frames. Our proposed method outperforms heatmap-based methods while utilizing significantly lower computation complexity and fewer model parameters. The FLOPs of our regression-based head are an almost negligible \textbf{1/7835} or \textbf{1/550} of those heatmap-based heads. We encourage our readers to refer to the supplementary materials for additional comparisons using smaller backbones, \textit{i.e.}, ResNet and MobileNet. The computational superiority of our proposed regression framework is of great value in the industry, particularly for real-time video applications. 

\textbf{Gains on Low-resolution Input.} 
In practical applications, especially on some edge devices with limited computation resources, it is common to use low-resolution images for reduced computational cost.  To explore the robustness of our model under different input resolutions, we compare our method with heatmap-based DCPose~\cite{DCPose_CVPR2021} on the PoseTrack2017 validation set. As shown in Table~\ref{tab:resolution_compare}, our method consistently outperforms DCPose across all input sizes. The results also show that the performance of heatmap-based methods decreases significantly with low-resolution input. For example, at an input resolution of $64 \times 64$, our proposed method outperforms DCPose by \textbf{20.3} mAP.


\subsection{Ablation Study}
We conduct ablation experiments to analyze the influence of each component using the PoseTrack2017 validation set. The temporal span $T$ is set to 1, consisting of one preceding and one subsequent frames, totalling 2 auxiliary frames, and the ResNet-50 backbone is employed.

\textbf{Impact of different modules.} 
Table~\ref{tab:impact_components} lists the performance impact of each module of our approach. When we adopt global pose features instead of modeling the temporal dependency at the joint level, \textit{i.e.}, without the JFD and STD modules, the algorithm achieves an mAP of 73.8, decreasing \textbf{4.8} points. When capturing spatiotemporal relations based on joints using Eq.~\ref{eqn:simple_dependency}, \textit{i.e.}, Space-Time coupling, the accuracy reaches \textbf{74.8} mAP. It aligns with our assumption that modeling temporal dependency at the joint level, as opposed to the entire pose, is more appropriate.
Furthermore, when using the SD and TD modules to separately model the temporal dynamic dependencies and spatial structure dependencies, the algorithm achieves the highest accuracy of \textbf{78.6} mAP. It proves that the temporal dependencies of each joint should be individually captured, as every joint exhibits an  independent temporal trajectory.
Meanwhile, we can see that the TD module capturing temporal dependencies has a much greater impact (\textbf{78.1}) on overall performance than the SD module capturing spatial dependencies (\textbf{71.4}). We believe this is mainly because the feature token extracted by the JFD module for each joint already contains its spatial context information within the pose. This means the spatial structure information complemented by the SD module is limited.

\textbf{Choice of JFD.} As discussed in Sec.~\ref{sec:JFD}, 
~\cite{Poseur_ECCV2022} proposes an alternative approach to constructing feature embeddings for each joint. We compare our JFD module with this method in Table~\ref{tab:jfd_choice}, where ~\cite{Poseur_ECCV2022} (b) augments the feature embedding of each joint with relevant contextual features while~\cite{Poseur_ECCV2022} (a) does not. As can be seen, our approach improves accuracy by \textbf{4.0} points with a relatively small computational overhead, achieving the highest accuracy.

\textbf{Auxiliary Frames.} 
In addition, we investigate the impact of using different numbers of auxiliary frames. 
The results presented in Table~\ref{tab:t_selection} consistently demonstrate that increasing the number of auxiliary frames leads to improved performance across various backbone networks.
It aligns with our intuition that more auxiliary frames can provide more complementary information, thereby facilitating the enhancement of pose estimation  for the key frame.

\section{Conclusion}

In this paper, we propose a novel and effective regression framework for video-based human pose estimation. Through the proposed Decoupled Space-Time Aggregation network (DSTA), we efficiently leverage temporal dependencies in video sequences for multi-frame human pose estimation, while reducing computational and storage requirements. Extensive experiments demonstrate the superiority of our approach over image-based regression methods as well as heatmap-based methods, opening up new possibilities for real-time video applications.

\section*{Acknowledgment} This work is supported by ``Pioneer'' and ``Leading Goose'' R\&D Program of Zhejiang Province (2024C01167).

{
    \small
    \bibliographystyle{ieeenat_fullname}
    \bibliography{main}
}

 \clearpage
\setcounter{page}{1}
\maketitlesupplementary

\section*{Appendix}
\label{sec:appendix}

In the supplementary material, we provide:\\\\
\noindent{\S}\textcolor[rgb]{1.00,0.00,0.00}{A} Additional Implementation Details. \\\\
\noindent{\S}\textcolor[rgb]{1.00,0.00,0.00}{B} Computation Complexity on More Backbones.\\\\
\noindent{\S}\textcolor[rgb]{1.00,0.00,0.00}{C} Experiments on PoseTrack2018/21 Datasets.\\\\
\noindent{\S}\textcolor[rgb]{1.00,0.00,0.00}{D} Additional Ablation Study. \\\\
\noindent{\S}\textcolor[rgb]{1.00,0.00,0.00}{E} Qualitative Results.

\section*{A. Additional Implementation Details}
\label{sec:appendix_impl_details}

\noindent{\textbf{Extracing Joint Embedding on HRNet.}} In the module of Joint-centric Feature Decoder (JFD), the feature embedding is extracted for each joint from the given global feature maps $\mathcal{F}_i(t')$ with $t'\in[t-T,t+T]$. 
In our implementation with HRNet backbones, specifically the HRNet-W48 variant, the high-resolution branch of the HRNet backbone is succeeded by a 1×1 convolutional layer (CONV) and a joint-wise fully connected feed-forward network (FFN) that encompasses a fully connected layer. The CONV layer consists of $n$ kernels, where $n$ represents the number of pose joints, resulting in a dedicated feature map for each joint. Subsequently, each individual feature map corresponding to a specific joint is flattened and fed into the shared FFN network, ultimately yielding its 32-bit feature embedding. \\\\
\noindent\textbf{Dataset.}
We evaluate our models on three widely-utilized video-based benchmarks for human pose estimation: PoseTrack2017~\cite{PoseTrack2017_CVPR2017}, PoseTrack2018~\cite{PoseTrack2018_CVPR2018}, and PoseTrack21~\cite{PoseTrack21_CVPR2022}. 
Specifically, PoseTrack2017 includes 250 video clips for training and 50 videos for validation , with a total of 80, 144 pose annotations. PoseTrack2018 considerably increases the number of clips, containing 593 videos for training, 170 videos for validation, and a total of 153, 615 pose annotations. Both datasets identify 15 keypoints, with an additional label for joint visibility. The training videos are densely annotated in the center 30 frames, and validation videos are additionally labeled every four frames. PoseTrack21 further enriches and refines PoseTrack2018 especially for annotations of small persons and persons in crowds, including 177, 164 human pose annotations. \\\\
\noindent\textbf{Optimization.} We incorporate data augmentation including random rotation [-45°, 45°], random scale [0.65, 1.35], truncation (half body), and flipping during training.  
We adopt the AdamW optimizer~\cite{Adam_ICTAI2019} to train the entire network for 40 epochs, with a base learning rate of 2\textit{e}-4, which is reduced by an order of magnitude at the 20\textit{th} and 30\textit{th} epochs. $\beta_1$ and $\beta_2$ are set to 0.9 and 0.999, respectively, and weight decay is set to 0.01. 

\begin{table}
\small
    \centering
    \setlength{\tabcolsep}{1mm}{
    \begin{tabular}{l|cccc}
       \multirow{2}{*}{Method}  &  \multirow{2}{*}{\#Params} &  GFLOPs   &  GFLOPs & \multirow{2}{*}{mAP} \\
       & & of Backbone & of Net. Head & \\
       \hline
       \multicolumn{5}{l}{\textit{heatmap-based}}   \\
       PoseWarper~\cite{PoseWarper_NIPS2019}  &  39.1M  &  4.1  &  90.3 & 75.9   \\
       DCPose~\cite{DCPose_CVPR2021}  & 35.6M   & 4.1  & 21.7 & 77.1 \\
       \hline
       \multicolumn{5}{l}{\textit{regression-based}} \\
       \textbf{DSTA (Ours)}  &  \textbf{24.6M}  &  4.1  & \textbf{0.01} & \textbf{78.6}  \\
    \end{tabular}}
    \caption{\textbf{Computation complexity} with ResNet-50 backbone. \#Params includes the parameters of entire network.  All methods utilize the same two auxiliary frames as  in~\cite{DCPose_CVPR2021}.}
    \label{tab:complexity_compare_resnet}
\end{table}

\begin{table}
\small
    \centering
    \setlength{\tabcolsep}{1mm}{
    \begin{tabular}{l|cccc}
       \multirow{2}{*}{Method}  &  \multirow{2}{*}{\#Params} &  GFLOPs  &  GFLOPs & \multirow{2}{*}{mAP} \\
       & & of Backbone & of Net. Head & \\
       \hline
       \multicolumn{5}{l}{\textit{heatmap-based}}   \\
       PoseWarper~\cite{PoseWarper_NIPS2019}  &  14.8M  &  0.35  &  73.5 & 67.7   \\
       DCPose~\cite{DCPose_CVPR2021}  & 11.3M   & 0.35  & 4.9 & 68.8 \\
       \hline
       \multicolumn{5}{l}{\textit{regression-based}} \\
       \textbf{DSTA (Ours)}  &  \textbf{2.4M}  &  0.35  & \textbf{0.01} & \textbf{71.0}  \\
    \end{tabular}}
    \caption{\textbf{Computation complexity} with MobileNet-V2 backbone. \#Params includes the parameters of entire network.  All methods utilize the same two auxiliary frames as  in~\cite{DCPose_CVPR2021}.}
    \label{tab:complexity_compare_mobilenet}
\end{table}

\section*{B. Computation Complexity on More Backbones}
\label{sec:appendix_computation complexity}
Tables~\ref{tab:complexity_compare_resnet} and ~\ref{tab:complexity_compare_mobilenet} present additional comparisons of computation complexity between our regression-based method and the heatmap-based methods. These experiments were conducted on the PoseTrack2017 validation set using the ResNet-50 and MobileNet-V2   backbones, respectively. 
While implementing the heatmap-based PoseWarper~\cite{PoseWarper_NIPS2019} and DCPose~\cite{DCPose_CVPR2021}, we utilized their official open-source codes.
In their network heads, similar to SimpleBase~\cite{SimplePose_ECCV2018}, we employed 3 deconvolution layers to generate high-resolution heatmaps from the backbones.

As shown, our method outperforms heatmap-based methods in both backbones, while utilizing significantly lower computation complexity and fewer model parameters. In addition, when compared to the HRNet backbone's results presented in Table~\ref{tab:complexity_compare} of the main paper, our method achieves even greater savings in computational costs and model parameters on these smaller backbone networks.
For instance, when utilizing the MobileNet-V2 backbone, our regression-based network incorporates a mere \textbf{2.4} million parameters, whereas the heatmap-based networks demand a significantly higher number, specifically 14.8 million and 11.3 million parameters. On the other hand, when employing the ResNet-50 backbone, the FLOPs of our regression-based head are almost negligible, accounting for just \textbf{1/9030} or \textbf{1/2170} of those required by the heatmap-based heads. The superior computational and storage efficiency of our proposed regression framework holds immense value in the industry, especially for edge devices and real-time video applications. 

\section*{C. Experiments on PoseTrack2018/21 Datasets}
Tables~\ref{tab:sota_compare_posetrack2018} and \ref{tab:sota_compare_posetrack21} present the comparisons of our method with the state-of-the-art methods on the PoseTrack2018 and PoseTrack21 valdation sets, respectively.
These results further demonstrate that our proposed regression-based method achieves performance that is either superior to, or at the very least, on par with the state-of-the-art heatmap-based methods. 

\begin{table}
\footnotesize
    \centering
    \setlength{\tabcolsep}{0.5mm}{
    \begin{tabular}{l|c|ccccccc|c}
       Method  & Bkbone & Head	 & Should. & Elbow	& Wrist & Hip & Knee & Ankle & \textbf{Mean} \\
       \hline
       \multicolumn{10}{l}{\textit{heatmap-based}}   \\
       \fontsize{6}{1}\selectfont AlphaPose~\cite{AlphaPose_ICCV2017} & \fontsize{6}{1}\selectfont ResNet-50 & 63.9 &	78.7 &	77.4 &	71.0 &	73.7 &	73.0 &	69.7 &	71.9  \\
       \fontsize{6}{1}\selectfont MDPN~\cite{MDPN_ECCV-W2018} & \fontsize{6}{1}\selectfont ResNet-152&	75.4 &	81.2 &	79.0 &	74.1 &	72.4 &	73.0 &	69.9 &	75.0 \\
       \fontsize{6}{1}\selectfont Dyn.-GNN~\cite{DynGNN_CVPR2021} & \fontsize{6}{1}\selectfont HRNet-W48&80.6 &	84.5 &	80.6 &	74.4 &	75.0 &	76.7 &	71.8 &	77.9 \\
       \fontsize{6}{1}\selectfont PoseWarp.~\cite{PoseWarper_NIPS2019} & \fontsize{6}{1}\selectfont HRNet-W48 &	79.9 &	86.3 &	82.4 &	77.5 &	79.8 &	78.8 &	73.2 &	79.7 \\
       \fontsize{6}{1}\selectfont PT-CPN++~\cite{PTCPN_ECCVW2018} &\fontsize{6}{1}\selectfont CPN~\cite{CPN_CVPR2018}&	82.4 &	\textbf{88.8} &	\textbf{86.2} &	\textbf{79.4} &	72.0 &	80.6 &	\textbf{76.2} &	80.9 \\
       \fontsize{6}{1}\selectfont DCPose~\cite{DCPose_CVPR2021} & \fontsize{6}{1}\selectfont HRNet-W48 &	84.0 &	86.6 &	82.7 &	78.0 &	80.4 &	79.3 &	73.8 &	80.9 \\
       \fontsize{6}{1}\selectfont DetTrack~\cite{DetTrack_cvpr2020} & \fontsize{6}{1}\selectfont HRNet-W48 &	84.9 &	87.4 &	84.8 &	79.2 &	77.6 &	79.7 &	75.3 &	81.5 \\
      \fontsize{6}{1}\selectfont FAMIPose~\cite{FAMIPose_CVPR2022} & \fontsize{6}{1}\selectfont HRNet-W48 &	85.5 &	87.7 &	84.2 &	79.2 &	81.4 &	\textbf{81.1} &	74.9 &	\textbf{82.2} \\
       \hline
       \multicolumn{10}{l}{\textit{regression-based}} \\
       \fontsize{6}{1}\selectfont \textbf{DSTA (Ours)} & \fontsize{6}{1}\selectfont ResNet-152  &	85.2 &	87.1 &	80.5 &	74.4 &	79.6 &	78.0 &	69.7 &	79.6  \\
      \fontsize{6}{1}\selectfont \textbf{DSTA (Ours)} &  \fontsize{6}{1}\selectfont HRNet-W48  &  \textbf{86.2} &	\textbf{88.6} &	84.2 &	78.5 &	\textbf{82.0} &	79.2 & 73.7 & \textbf{82.1} \\
      
       \cellcolor{light-gray}\fontsize{6}{1}\selectfont \textbf{DSTA (Ours)} & \cellcolor{light-gray}\fontsize{6}{1}\selectfont ViT-H  &	\cellcolor{light-gray}85.9 &	\cellcolor{light-gray}88.8 &	\cellcolor{light-gray}85.0 &	\cellcolor{light-gray}81.1 &	\cellcolor{light-gray}81.5 &	\cellcolor{light-gray}83.0 & \cellcolor{light-gray}77.4 &	\cellcolor{light-gray}83.4 \\
    \end{tabular}}
    \caption{\textbf{Comparison with the SOTA} on PoseTrack2018 val. set. Similar to FAMI-Pose~\cite{FAMIPose_CVPR2022}, our proposed DSTA sets the temporal span $T$ to 2, consisting of two preceding and two subsequent frames, totalling four auxiliary frames.}
    \label{tab:sota_compare_posetrack2018}
\end{table}

\begin{table}
\footnotesize
    \centering
    \setlength{\tabcolsep}{0.5mm}{
    \begin{tabular}{l|c|ccccccc|c}
       Method  & Bkbone & Head	 & Should. & Elbow	& Wrist & Hip & Knee & Ankle & \textbf{Mean} \\
       \hline
       \multicolumn{10}{l}{\textit{heatmap-based}}   \\
       \fontsize{6}{1}\selectfont SimBase.~\cite{SimplePose_ECCV2018} & \fontsize{6}{1}\selectfont ResNet-152 & 	80.5 &	81.2 &	73.2 &	64.8 &	73.9 &	72.7 &	67.7 &	73.9 \\
       \fontsize{6}{1}\selectfont HRNet~\cite{HRNet_CVPR2019} & \fontsize{6}{1}\selectfont HRNet-W48 &	81.5 &	83.2 &	81.1 &	75.4 &	79.2 &	77.8 &	71.9 &	78.8 \\
       \fontsize{6}{1}\selectfont PoseWarp.~\cite{PoseWarper_NIPS2019} & \fontsize{6}{1}\selectfont HRNet-W48 &82.3 &	84.0 &82.2 &	75.5& 	80.7 &	78.7 &	71.6 &	79.5 \\
       \fontsize{6}{1}\selectfont DCPose~\cite{DCPose_CVPR2021} & \fontsize{6}{1}\selectfont HRNet-W48 &	83.7 &	84.4 &	82.6 &	\textbf{78.7} &	80.1 &	79.8 &	74.4 &	80.7 \\
      \fontsize{6}{1}\selectfont FAMIPose~\cite{FAMIPose_CVPR2022} & \fontsize{6}{1}\selectfont HRNet-W48 &	83.3 &	85.4 &	82.9 &	78.6 &	81.3 &	\textbf{80.5} &	\textbf{75.3} &	81.2 \\
       \hline
       \multicolumn{10}{l}{\textit{regression-based}} \\
       \fontsize{6}{1}\selectfont \textbf{DSTA (Ours)} & \fontsize{6}{1}\selectfont ResNet-152  &	86.1 &	85.5 &	80.0 &	74.6 &	80.5 &	76.9 &	70.2 &	79.6  \\
      \fontsize{6}{1}\selectfont \textbf{DSTA (Ours)} &  \fontsize{6}{1}\selectfont HRNet-W48  &  \textbf{87.5} &	\textbf{86.6} &	\textbf{83.3} &	\textbf{78.7} &	\textbf{82.7} &	78.3 &	73.9 &	\textbf{82.0} \\
      
       \cellcolor{light-gray}\fontsize{6}{1}\selectfont \textbf{DSTA (Ours)} & \cellcolor{light-gray}\fontsize{6}{1}\selectfont ViT-H  &	\cellcolor{light-gray}87.5 &	\cellcolor{light-gray}87.0 &	\cellcolor{light-gray}84.2 &	\cellcolor{light-gray}81.4 &	\cellcolor{light-gray}82.3 &	\cellcolor{light-gray}82.5 &	\cellcolor{light-gray}77.7 &	\cellcolor{light-gray}83.5 \\
    \end{tabular}}
    \caption{\textbf{Comparison with the SOTA} on PoseTrack21 val. set. Similar to FAMI-Pose~\cite{FAMIPose_CVPR2022}, our proposed DSTA sets the temporal span $T$ to 2, consisting of two preceding and two subsequent frames, totalling four auxiliary frames.}
    \label{tab:sota_compare_posetrack21}
\end{table}

\section*{D. Additional Ablation Study}
\textbf{Size of Joint Tokens.} In this additional study, we conduct experiemnts to examine the influence of the adopted size of the joints' feature embedding (\textit{i.e.}, joint token). Table~\ref{tab:tokensize} presents the performance variations resulting from different joint token sizes on the PoseTrack2017 validation set. As the size of the joint tokens increases, a gradual improvement in performance can be observed. However, beyond a size of 16, the performance tends to plateau, suggesting that further increases in token size do not yield commensurate improvements.
This indicates that each pose joint requires a sufficiently large feature token to store its relevant feature information, but a too large feature token will only cause spatial redundancy. Therefore, in our experiments, we have opted to use a token size of 32, striking a balance between capturing sufficient feature information and avoiding unnecessary spatial redundancy.

\begin{table}
    \centering
    \begin{tabular}{c|cccc}
       \#Token Size  & 8 & 16 & 32 & 64  \\
       \hline
       mAP &  76.1 & 78.0 & 78.6 & 78.6 \\
    \end{tabular}
    \caption{\textbf{Different sizes of joint tokens}. In the experimental setup, we utilized the ResNet-50 backbone along with two auxiliary frames.}
    \label{tab:tokensize}
\end{table}

\section*{E. Qualitative Results}
\begin{figure}
    \centering
    \includegraphics[width=1.0\linewidth]{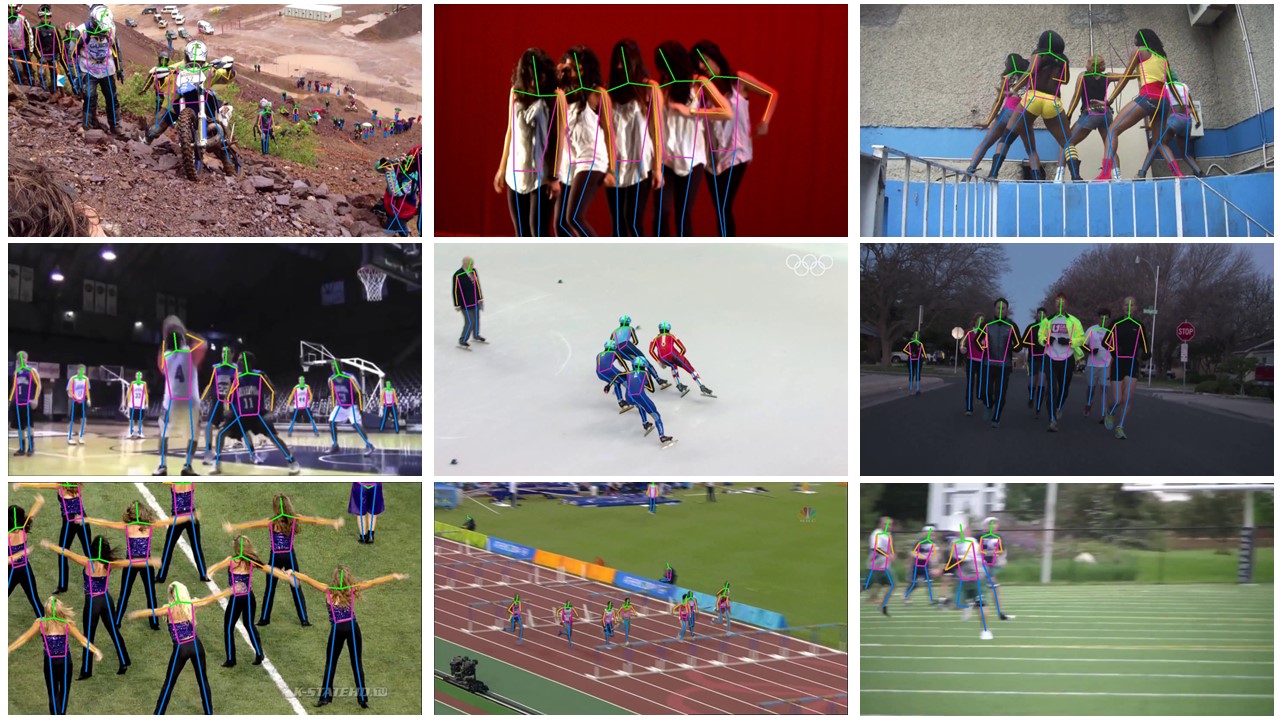}
    \caption{\textbf{Additional qualitative results} of our DSTA on the PoseTrack datasets.}
    \label{fig:qualitative_result_supplementary}
\end{figure}

Additional qualitative results on PoseTrack datasets are shown in Fig.~\ref{fig:qualitative_result_supplementary}. Additional results can be found in the accompanying video material.

\end{document}